# ATDN vSLAM: An All-Through Deep Learning-Based Solution for Visual Simultaneous Localization and Mapping

Mátyás Szántó[1*], György Richárd Bogár[1], László Vajta[1]

[1] Department of Control Engineering and Information Technology, Faculty of Electrical Engineering and Informatics, Budapest University of Technology and Economics, 2 Magyar tudosok Blvd., H-1117 Budapest, Hungary
* Corresponding author, e-mail: mszanto@iit.bme.hu

**Abstract**
In this paper, a novel solution is introduced for visual Simultaneous Localization and Mapping (vSLAM) that is built up of Deep Learning components. The proposed architecture is a highly modular framework in which each component offers state of the art results in their respective fields of vision-based Deep Learning solutions. The paper shows that with the synergic integration of these individual building blocks, a functioning and efficient all-through deep neural (ATDN) vSLAM system can be created. The Embedding Distance Loss function is introduced and using it the ATDN architecture is trained. The resulting system managed to achieve 4.4% translation and 0.0176 deg/m rotational error on a subset of the KITTI dataset. The proposed architecture can be used for efficient and low-latency autonomous driving (AD) aiding database creation as well as a basis for autonomous vehicle (AV) control.

## 1 Introduction

Mobile robotics is a field that has seen ongoing scrutiny in the past several decades. A few of the most important tasks to be solved during the operation of autonomous robotic agents is positioning and understanding their environments as well as navigation. Simultaneous Localization and Mapping (SLAM) attempts to solve the above functions in a concurrent manner [1, 2].

With the appearance of the endeavor for modernizing traffic automation and the creation of self-driving passenger and transportation vehicles, SLAM research has surged into the spotlight in the past few years [3, 4]. A crowdsourcing-based mapping solution, such as CrowdMapping [5], whose purpose is to offer low latency and accurate map updates for autonomous vehicles, can be supported and aided using robust SLAM technologies. One well-researched subfield of SLAM is visual SLAM, or vSLAM [6–8] for short, wherein localization, mapping and sometimes path planning is achieved using visual information. That information may be supplied in the form of strictly 2D images – e.g., passive monocular camera RGB image streams or stereo camera systems – or 3D image information provided by specific sensors – e.g., light detection and ranging (LiDAR) sensors or active RGB + Depth (RGB-D) cameras. The visual information is often supplemented with data prescribing the motion of the robot or agent carrying the image acquisition device – for example via an inertial measurement unit (IMU).

Monocular, passive vSLAM methods can be categorized based on the number of points identified and analyzed in the environment [7]. A number of methods are classified as feature-based [9–11], i.e. the resulting maps are based on a relatively small number of feature points in the depicted environment. Another category is direct [12–15], wherein the results are calculated based on information available for each pixel of the input image. As part of direct methods, semi-dense and hybrid [16, 17] techniques have also been gaining attention: these solution types combine the advantages of the two previously mentioned solution types.

The above methods differ in their ways of reaching a result; however, they share the same foundations, in that they are all monocular visual simultaneous localization and mapping techniques. These methods usually have a pipeline structure of the following components:
- initialization;
- visual odometry and pose estimation;
- mapping;
- map refinement via new information.

As a result of the currently observable immense progress of the scientific fields surrounding machine learning, and particularly Deep Learning (DL), many approaches have appeared that are able to supply solutions for each of the above listed sub-tasks of a vSLAM framework. However, to the best of our knowledge, no previous research paper has been published that offers an all-through DL pipeline. Such a pipeline combines the advances that novel machine learning methods offer in contrast to traditional, hand-crafted algorithms, while also offering close to state-of-the-art results for the measurable subtasks of dense vSLAM methods.

Our contributions in this paper are as follows:
- we present an all-through Deep Learning-based pipeline for direct monocular dense vSLAM calculation that comprises of a visual odometry, and a map encoding component (Fig. 1);
- we introduce the step-by-step and overall training processes for this framework – we experiment with a novel Variational UNet core module;
- we propose a novel loss function that we call Embedding Distance Loss (EDL). We use this function to train the mapping component of our system;
- we show our results achieved on common benchmarks and compare them to the state-of-the-art.

This paper is constructed as follows: in Section 2, we describe related works and technical literature, and then in Section 3, we introduce our methods and the main considerations behind our work. Then in Section 4, we present our results and compare them to state-of-the-art methods. In Section 5, we discuss our results, and then conclude our paper in Section 6.

## 2 Related work
In Section 2, we briefly summarize the technical results and other seminal works that are related to our scientific contributions described in this paper.

### 2.1 Simultaneous Localization and Mapping
Simultaneous Localization and Mapping methods – SLAM for short – have been a thoroughly researched area in computer science and in mobile robotics for the past few decades. The basis of the technology concerns itself with a topic that is of utmost importance in robotics, namely navigating the robot in a previously unknown environment using data that is collected by the robot during its operation [1, 2, 18, 19]. The navigation requires the robot to understand and chart its surroundings while localizing itself in the environment and thus constantly gaining information and expanding its knowledgebase.

Although, early solutions of SLAM had utilized traditional robotics sensor elements, such as IMU, and versatile distance and waypoint detectors, with more recent advances in the field of visual informatics and machine vision, another subgenre – often referred to as visual SLAM or vSLAM – has been established [7]. Modern vision-based SLAM methods can be categorized into three classes based on the data obtained by the sensors: visual-only SLAM (see Sub-subsection 2.1.1), Visual-Inertial SLAM or VI-SLAM [20–22], and RGB-Depth SLAM or RGB-D SLAM [23–25].

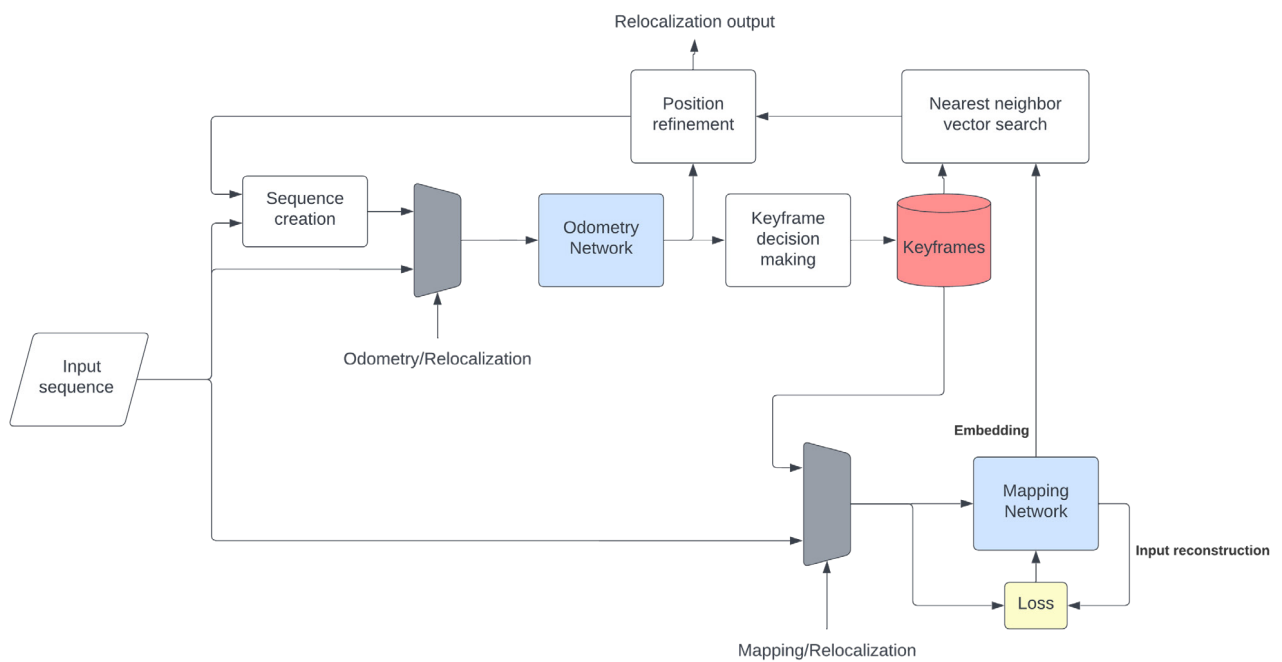

**Fig. 1** Our all-through deep neural (ATDN) vSLAM architecture

Since the solution introduced in this paper is based solely on monocular visual information, the following parts of Section 2 present a more in-depth review of visual-only SLAM methods with particular emphasis put on techniques utilizing Deep Neural Networks (DNNs).

### 2.1.1 Feature-based and direct monocular vSLAM

Because of the low cost and high availability of regular RGB cameras, visual-only SLAM solutions have been well-researched in the past years. Based on the number of cameras used, there are different solutions in the literature using monocular and stereo camera systems. In this paper, we are describing our method that only makes use of monocular RGB cameras. Monocular visual-only SLAM methods can be partitioned into two main categories:

*Feature-based methods* rely on a certain number of keypoints – or points of interest – extracted from the input images. Based on feature points appearing on consecutive images, the motion of the camera as well as the sparse environment can be calculated via matching the descriptors of these frames. Davison et al. [9] proposed the first truly monocular vSLAM method using keypoints and an extended Kalman filter. The usage of keyframes and bundle adjustment for a feature-based solution was introduced by Klein and Murray [11]. ORB-SLAM2 [26], the successor of ORB-SLAM [10], is considered the state-of-the-art monocular feature-based vSLAM method. This technique introduces parallel calculation of tracking, local mapping, and loop closure via bundle adjustment.

*Direct methods* use all pixels of the input images to compare them and calculate the photometric error. The environment mapping is based on the minimized error, resulting in a dense, semi-dense, or sparse reconstruction. Dense Tracking And Mapping (DTAM) [12], which is based on PTAM [11], is the first direct method proposed, combining the tracking portion of PTAM with a dense tracking step that enables the calculation of depth maps. As the name suggests, DTAM produces dense reconstructions. DTAM was advanced and optimized by Ondruska et al. [27] to use GPU acceleration for mobile devices. Engel et al. [13] introduced a semi-dense reconstruction method in large-scale direct SLAM (LSD-SLAM), where tracking and keyframe selection is based on motion stereo constraints. The direct sparse odometry (DSO) [15] technique introduces a sparse direct reconstruction method for mapping. Gao et al. [28] advanced DSO with loop closure and pose-graph optimization. Tateno et al. introduced CNN-SLAM [14], a dense direct vSLAM solution that made use of convolutional neural networks for depth estimation, semantic segmentation, and pose-graph optimization. It was the first work to gain international attention that used Deep Learning in its core functionality.

### 2.2 Deep Learning and vSLAM

Neural networks can be applied in many different parts of a vSLAM framework. These approaches can be categorized according to the different tasks in a vSLAM method [29].

### 2.2.1 Visual Odometry

Visual Odometry (VO) [30] is one of the most essential components of a vSLAM pipeline. As described in previous works [31], vSLAM pipelines are combinations of VO and global optimization steps.

For visual odometry, Deep Learning can be applied in an end-to-end manner or in a hybrid way. We can differentiate between supervised and unsupervised Deep Learning methodologies in general, a categorization that also holds for visual odometry. Previous results [14], and the works disussed below [32–35] managed to show that scale-correct results can be assured through supervised learning – in contrast with the traditional VO methods, which did not manage to solve scale-ambiguity without the introduction of various scale-defining inputs, such as dimensions of known objects in an image. Training a neural network for the solution of VO given a metrically labeled training dataset also solves the problems of scale drift and pure rotation.

Wang et al. [32] showed that through training a VO network in the supervised manner, it is possible to achieve real world metric results if the ground truth data is also measured in a real-world reference. Another key solution is presented in [33]: CLVO introduces curriculum learning with a windowed composition layer which ensures the long-term consistency of odometry estimations.

A possible solution for unsupervised learning VO, is to simultaneously learn and optimize depth estimation with odometry. This can cause stronger correlation with the 3D reconstruction, for example with a point-cloud based solution [34].

Currently, state-of-the-art results can also be achieved with the so-called hybrid approaches which take the best qualities of the classical methodologies and Deep Learning solutions.

A good example is the D3VO architecture [35], which achieves high performance in the visual odometry field. Although hybrid solutions can currently outperform fully learned odometry, the good separation of the two approaches and the choice of algorithms can thoroughly affect the results.

**2.2.2 Mapping**
Besides odometry, another task in the vSLAM pipeline is mapping. Mapping is the process when the explored environment is converted into a format that is easy to store and to reuse later for relocalization and optimization. Mapping methods can be distinguished by the representation of the created map. One of the most straightforward ways is to store and process the environment using a 3D geometry representation. While some 3D representations can be acceptable for SLAM purposes, others have more limitations in the field.

Voxel representation is the 3D equivalent of pixels; therefore, we can approach the processing of these structures similarly to the image processing field. It is also true for Deep Learning: convolutional kernels for pixels have their equivalents for voxels. One limitation of 3D filtering and convolution is in the number of optimizable parameters: while the number of weights in a 2D image convolution kernel is usually the square of the kernel width (with the exception of non-quadratic kernels), the voxel kernel scales with the third power of the dimensions. This makes the relatively small kernel size 3 (width, height, and depth alike) which creates 9 weights in a 2D convolution, a mid-sized layer with 27 weights in the 3D space. On the other hand, it should be considered that the activations of voxel convolution must be calculated in an additional dimension, which also greatly increases computational needs. While this representation has the advantage of having exact neighboring points, this also brings a disadvantage, namely that no empty space can be eliminated for computation and memory optimization.

Another option for representing 3D geometry is the point cloud (PC) representation. A PC is made of a set of 3D points in space, usually with location and color information. PCs give a lightweight representation of 3D, leaving out empty spaces, however this brings the disadvantage of no obvious neighboring points. In the field of point clouds, Deep Learning is a limited methodology for there is no order or explicit connection between the points. Despite the limitations, there are a few approaches aiming at achieving 3D processing with point clouds [12, 36].

Mesh representation of 3D can be familiar from the computer graphics field; however, it is worth the investigation as a possible candidate of vSLAM mapping format. Mesh representation can propose the advantage of point clouds having no unnecessary empty spaces stored and the advantage of the voxels having exact neighboring points inside an object. With these positive attributes, mesh representation can be an ideal choice for experimental Deep Learning-based mapping and map processing methods.

Another possibility for scene understanding is using semantic information. Semantic segmentation is one of the main research fields in computer vision and Deep Learning. Numerous solutions are available in the literature with constantly increasing performance [36].

A new mapping solution came with the introduction of neural networks in the SLAM pipeline. General Map is a special type, which is represented by the output of a network. A general map does not explicitly represent any of the previously mentioned 3D or semantic constructions, as it is an embedding of the explored environment.

The most straightforward way to generate a general map is the usage of autoencoders. These systems can reduce the dimensionality of the input data with minimal loss of representative data. The code that represents a view (an image) in the environment is the embedding, which is generated by the bottleneck layer of the autoencoder.

With this approach, the weights of the trained network can be interpreted as the map of the exploration. Learning the latent space (the space of possible codes generated by the encoder part) can be useful for later use of the latent space vector – e.g., for relocalization purposes.

However, basic autoencoders are not bound to generate a continuous latent space. This problem can be solved with the application of a variational autoencoder. According to experimental results, general maps (the complexity of the encoding network) scale better than linear with increasing explorable environment [37]. This can be a significant advantage for a solution, where map data transfer speed is critical – e.g., in an automotive environment.

Global localization is the task during which the SLAM algorithm associates global coordinates of the camera pose from one or more input images. Localization is dependent on the mapping methodology. Using a general map, two options are possible: the embedding difference comparison or a further neural network for global pose regression.

Besides the above-mentioned building blocks, a SLAM system can contain other optimization algorithms that can further improve the consistency of the output of the separate tasks.

**2.2.3 vSLAM methods with Deep Learning components**
One SLAM solution that utilizes Deep Learning is CNN-SLAM [14]; In this work, neural networks are used to predict depth and semantic labels from input images. The depth estimation calculated by the convolutional network is fused with traditional depth prediction: the neural network-based depth is calculated for the keyframes and is refined by small baseline stereo depth estimation.

CodeSLAM [38] is another neural network-based SLAM solution, which takes advantage of Deep Learning mainly in the mapping process. The solution proposes a compact scene geometry representation by conditioning depth for an autoencoder with image intensity. This proposed SLAM method can be used in a keyframe-based solution, where newly registered keyframes have a code representing the local scene geometry. The code is only required to represent those geometric properties, which cannot be predicted from the input image.

**3 Methods**
In Section 3, the basic methods behind our proposed solution are introduced, including the all-through Deep Learning vSLAM pipeline, the novel Embedding Distance Loss (EDL) function and the individual Deep Learning-based components together with their training methods.

**3.1 All-through deep neural monocular vSLAM pipeline**
The proposed network architecture (Fig. 2) solves two of the main tasks with the use of deep neural networks. Visual Odometry tends to provide increasingly accurate results in mid-range scenarios like the one presented by the KITTI odometry dataset sequences. The second main field is the mapping process, during which we encode the keyframes of the explored environment in a way that other images, which did not take part in the training process, could be recognized. For this task, autoencoders and other similar encoding architectures are in focus. Variational autoencoders can be useful, because of the continuous latent space they generate for the encoded vectors. However, not strictly latent spaces have the potential to be usable in a vSLAM mapping, wherein the distances of the generated latent space vectors can represent the distances of the actual codes. For this purpose, a loss component can be introduced. The Embedding Distance Loss (EDL) component is formulated as follows:

$$\text{EDL} = \left| \frac{\mathbf{E}_i - \mathbf{E}_{i+12}}{\mathbf{E}_{i+2} - \mathbf{E}_{i+12}} \right|^2 - \left| \frac{p_i - p_{i+12}}{p_{i+2} - p_{i+12}} \right|^2, \quad (1)$$

where $p_i$ represents the ground truth (or the odometry based estimation) for the camera positions of the corresponding input image, and $\mathbf{E}_i$ is the generated latent space vector.

EDL measures the difference in distance rates instead of direct distances. The advantage of this approach is that it does not force the network to generate codes with the exact same distances, only the rate of the differences is important.

Another advantage is that using only this loss term can eliminate the need of an autoencoder-like structure, therefore eliminating the residual error that a decoder generates while trying to reconstruct a high-resolution input image. By using EDL, the network can generate codes without any limitation on their magnitude. The hardship lies in the application of this function: at least three latent space vectors and points are needed.

The Embedding Distance Loss component can be used to force the autoencoder to generate the latent space vectors of the separate images having distances proportional with the difference of the camera positions. Further experiments are required to inspect whether an initial odometry estimate like the output of the proposed network is accurate enough to gain vector distances useful for robust loop closure detection. Using only EDL could not cause robust convergence in the network training, however it can be useful if the initial odometry estimate is accurate enough.

For our initial experiments, a CNN based encoder was used with only the ED loss component and the images were given in the batch dimension. Results with this experimental setup showed minimal or no convergence. A reason of this phenomenon is that the network is unable to extract embeddings with only the differences as an objective.

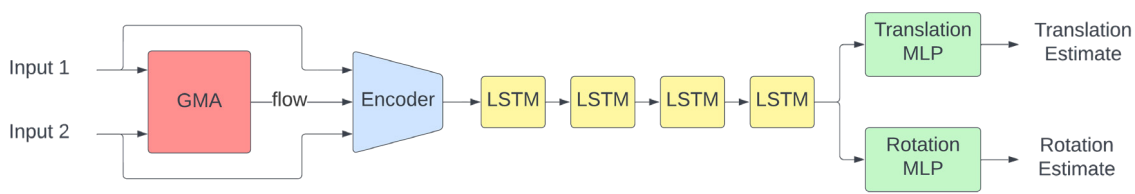

**Fig. 2** Network architecture used for visual odometry and mapping

One possible solution can be reorganizing the network to extract the differences between the latent space vectors. This can be achieved with an LSTM, a transformer or with the images concatenated in the channel dimension. The other possible use of the loss is to use it as a secondary regularization component in the training of an autoencoder-like structure.

Besides odometry and mapping, global localization remains the third core functionality that the proposed vSLAM system aims to solve. While in this system the odometry works together with the mapping in a loosely coupled way, the localization task can be solved both in a loosely and a more tightly coupled way.

The loosely coupled solution takes advantage of the distances of the latent space vectors. After the mapping network is trained, embeddings are computed for all keyframes. With the calculated embedding the relocalization process took three steps. First, the embedding has to be computed of the image to be localized. Second, the magnitude of difference must be determined between the new embedding and all stored keyframe embeddings. The keyframe with minimal difference can be determined as the closest keyframe to the current position. To further optimize relocalization precision, an odometry estimation can be calculated between the estimated keyframe and the new frame.

Our experimental results shown in Section 4 were achieved using the loosely coupled method.

The tightly coupled alternative for relocalization is the construction of a dense neural network that estimates position and orientation directly from mapping embeddings.

The remaining framework focuses on the orchestration and synchronization of these core elements and managing further optimizations.

### 3.2 Datasets
There are several datasets commonly used in training, validating, and evaluating vision-based autonomous vehicle-related technologies, such as vSLAM methods. These include KITTI [39], KITTI360 [40], The Oxford RobotCar Dataset [41], and recently the ApolloScape Dataset [42], Waymo Dataset [43]. One of the most used datasets, however, is still the KITTI [39] dataset that enables benchmarking individual solutions, too.

Many novel methods publish results achieved on this latter database. Therefore, during our research, we mainly used this dataset for training as well as for testing and comparison of our solution with the state-of-the-art.

### 3.3 Network training
Odometry training was mainly done on the KITTI odometry dataset 00 sequence. To take advantage of the transfer learning based optical flow and make the training faster, optical flow values are precomputed and stored, therefore memory and processing time is reduced as GMA is not used during training time. As mentioned above, the CLVO paper introduces curriculum learning with a windowed composition layer. Our experiments showed that increasing sequence length results in better performance than the scaling of the composition loss. For training, the AdamW optimizer was used with a cosine annealing learning rate (lr) schedule starting at $lr = 10^{-3}$ and with a final rate of $lr = 10^{-6}$. For the training, we used an Nvidia Titan X GPU with 12 GB memory and coupled with an Intel i5 8600 K 2.8 GHz CPU. With this hardware setup, the average epoch time was 28.5 minutes. One stage consists of 5 to 10 epochs: the first two stages ($\alpha = 1$ and $\alpha = 0.7$) were 5 epochs, the third and fourth ($\alpha = 0.3$) were 10 epochs. The learning rate schedule restarted for every stage. The α parameter denotes the long-term consistency component of the loss function applied for curriculum learning as used by [33].

The mapping network is an autoencoder having a bottleneck with variational ability. The variational part can be turned on and off. If turned off, only the mean is given to the decoder. The network was trained using the same machine on the chosen keyframes for 10 epochs with a batch size of 8. The Adam optimizer is used with cosine annealing learning rate scheduler starting with $lr = 10^{-3}$ and ending with $lr = 10^{-5}$.

#### 3.3.1 Depth estimation
Depth estimation can play a supporting role in the ATDN vSLAM architecture. The current structure can benefit from depth information in the neural network-based blocks: the odometry and the localization. In both cases, depth can be added as a separate channel for the input convolution and the network can learn to pay more attention to the parts of the input images which represent a part of the environment that is closer to the camera. In case of the autoencoder, depth can also be used as a regularization component, therefore closer parts are meant to be reconstructed more precisely. However, in this work, we opted not to use depth estimation as part of our pipeline shown in Fig. 1, as it did not show notable improvements in the results.

#### 3.3.2 VAE
Variational autoencoders (VAE) are a subset of autoencoder architectures. Autoencoders are usually used for supervised learning scenarios, such as data complexity reduction. Generally, an autoencoder is built of three main parts: an encoder, a bottleneck layer, and a decoder. The decoder and the bottleneck layer generate the latent space vector which is the reduced representation of the input data. The decoder part is necessary for the supervised training: it aims to reconstruct the input and therefore prove that the latent space vector is a valid representation of the input data. The main advantage of the variational autoencoder in comparison with a basic autoencoder is that a VAE system generates continuous latent space vector which can be useful for the relocalization segment of a vSLAM architecture.

### 3.3.3 VO

In case of visual odometry, a CLVO-based solution is implemented and used with the pretrained GMA optical [44, 45] flow estimator network. With this solution, learning optical flow estimation can be a pretrained block of a transfer learning process. To eliminate most of the unnecessary computations, and reduce training time, optical flow values are computed as a dataset generation process and are saved in the tensor file representation of the used library for reusability. With this, only one-time optical flow calculation is needed for the training. Of course, for odometry inference, the GMA flow block is needed as an active part of the odometry estimator structure (Fig. 2).

## 4 Results

In Section 4, the results of the odometry and mapping part of our solution are presented. In Fig. 3, the X, Y, Z propagation of the data acquisition vehicle can be seen over the images taken from the KITTI 00 sequence. The estimations provided by our solution are close to the ground truth data with regard to the values provided for X and Z values (top and bottom graphs in Fig. 3, respectively), however, the Y estimate (middle graph in Fig. 3) gains a more significant difference from the training data. This deviation has a noteworthy impact on the results of the quantitative evaluation – shown in Table 1.

Fig. 4 provides an easily interpretable result of the odometry output of our solution, as it shows the actual route of the data acquisition vehicle (denoted as gt for ground truth) and the predicted route given as the output of our solution (denoted as pred for prediction). This representation provides a birds-eye-view map of the route, which was driven and used for sequence 00 of the KITTI dataset. We have also evaluated our solution for relocalization on the same sequence. For this experiment, a random keyframe was selected from the sequence, whose embedding was then calculated. Using the embedding distance calculation, similarity of the embedding of the selected keyframe, and embeddings of the keyframes were calculated. The results of this similarity evaluation can be seen in the top diagram of Fig. 5.

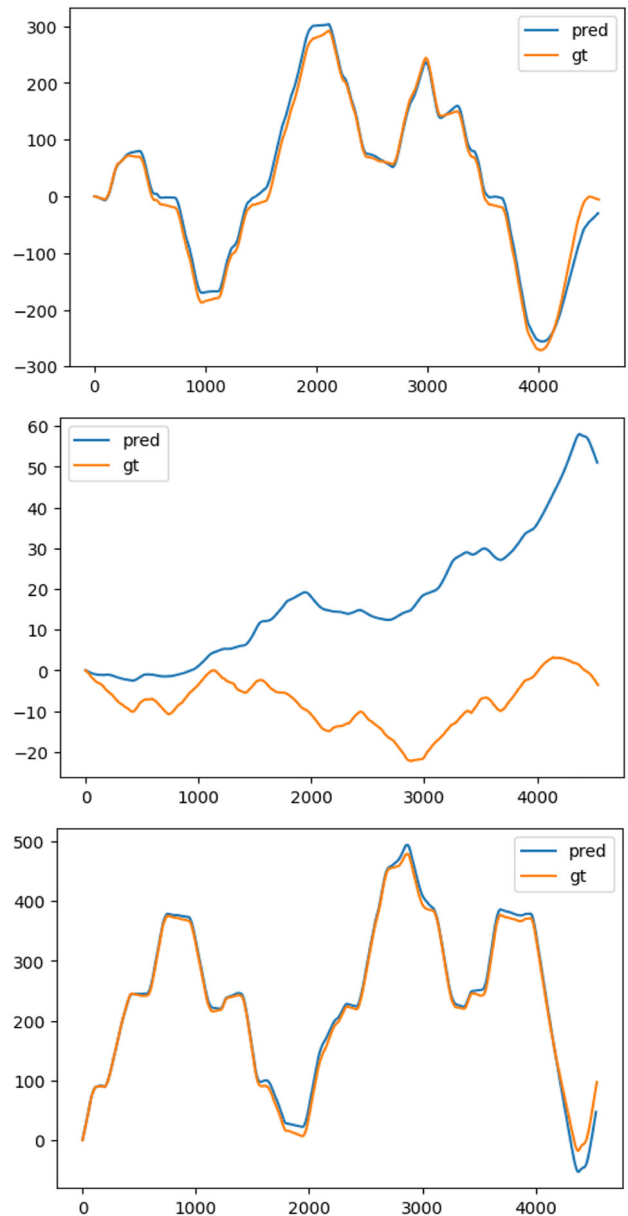

**Fig. 3** Results of mapping: propagation of X (top), Y (middle), and Z (bottom) coordinate over time on the KITTI 00 sequence – "gt" denotes ground truth, while "pred" denotes the output of our solution. Dimension of the vertical axes is metres.

The bottom illustration of Fig. 5 shows the histogram of these embedding distances. When a keyframe is chosen as the test frame, the resulting embedding distance is 0, validating the correct functionality of our solution.

Table 1 Comparing the results of our methods with state-of-the-art monocular RGB odometry solutions from the KITTI benchmark [39] (↓ denotes that lower numbers are better). The runtime results of our solution are provided with and without online optical flow calculation. Traditional feature point extraction SLAM methods are denoted with an asterisk (*).

| Method | Error | | Runtime | Environment |
|---|---|---|---|---|
| | Translation [%] ↓ | Rotation [deg/m] ↓ | [s] ↓ | |
| ORB-SLAM 2* | 1.15 | 0.0027 | 0.06 | 2 CPU cores @ >3.5 GHz |
| ESVO* | 1.42 | 0.0048 | 1 | 1 CPU core @ 2.5 GHz |
| D3VO | **0.88** | **0.0021** | 0.1 | 1 CPU core @ 2.5 GHz |
| GenPa-SLAM | 3.48 | 0.121 | 0.1 | GPU @ 2.5 GHz |
| Deep-AVO | 4.1 | 0.0125 | 0.01 | GPU @ 3.0 GHz |
| CUDA-Ego-Motion | 4.36 | 0.0052 | 0.001 | GPU @ 2.5 GHz |
| D3DLO | 5.4 | 0.0154 | 0.1 | GPU @ 2.5 GHz |
| Ours (w/ OF) | 4.4016 | 0.0176 | 0.27 | GPU @ 2.8 GHz |
| Ours (w/o OF) | | | **0.006** | |

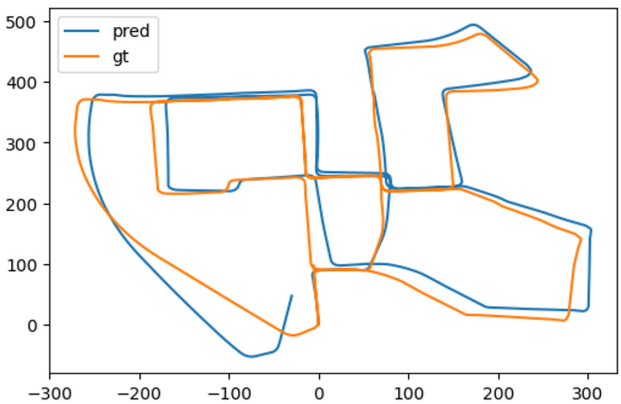

Fig. 4 Results of mapping: X-Z map of the KITTI 00 sequence - "gt" denotes ground truth, while "pred" denotes the output of our solution.

Fig. 6 shows the embedding distances and the distance histogram of a randomly chosen frame – which was not a keyframe, thereby wasn't included in the training process of the mapping network. For the relocalization of non-keyframe input images, the minimum distance has to be determined to find the matching location on the learned map.

There are some cases where it is a more difficult task, because there may be several similar keyframes with low embedding distances from the selected frame. However, it is visible on the histogram shown in Fig. 6, that the lowest embedding distance is still considerably lower than the mean distance of all the keyframes. An outlier search therefore is able to provide us with the best candidates, out of which a recursive match-search can yield the closest keyframe for relocalization.

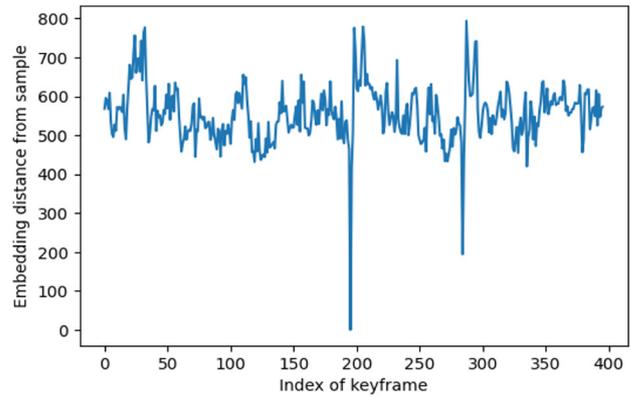

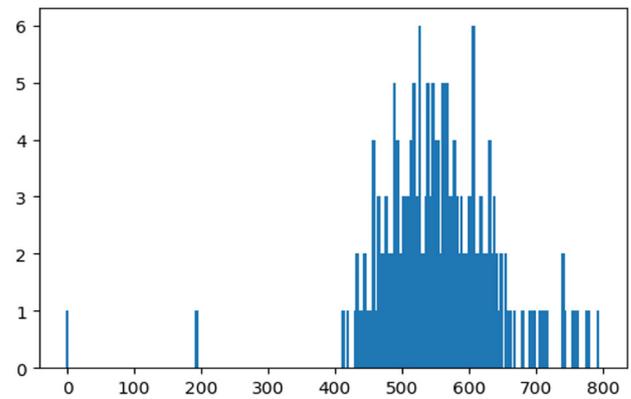

Fig. 5 Results of relocalization: embedding distances between embedding of a randomly selected keyframe (frame index: 2372) and all keyframes – as a function of the keyframe index (top) and the distribution shown as a histogram (bottom).

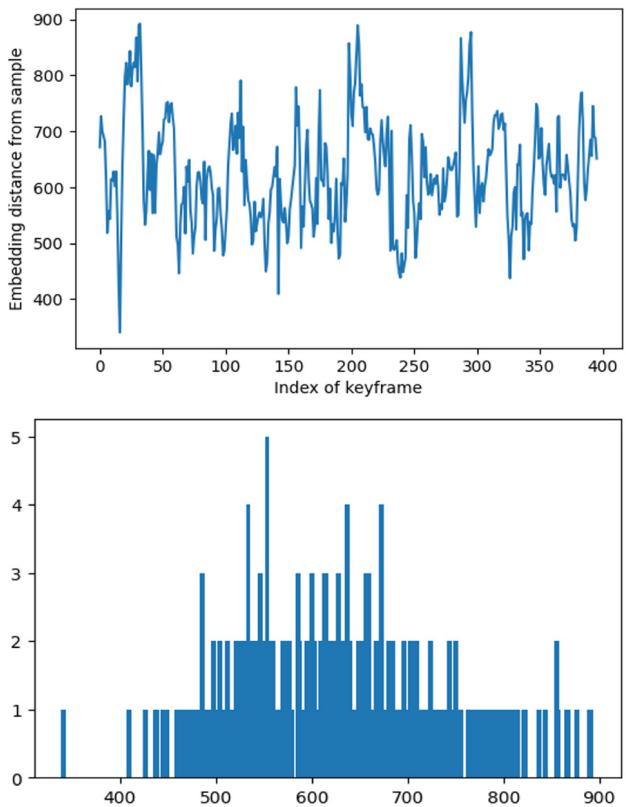

**Fig. 6** Results of relocalization: embedding distances between embedding of a randomly selected frame (frame index: 195) and all keyframes – as a function of the keyframe index (top) and the distribution shown as a histogram (bottom).

The results of our method are compared to the state-of-the-art (SOTA) shown in Table 1. It can be seen from Table 1, that while the error of our solution is en par with some SOTA solutions, its structure and high modularity might result in it achieving a considerably lower runtime. Our results come from the experimental training, which was only evaluated on the 00 sequence of the KITTI dataset.

Results of other solutions are from the official KITTI benchmark which were evaluated on the test sequences. For our solution, a significant performance drop can be expected for other sequences. To solve this, generalization is required: one solution candidate is training our system on a larger and more diverse set of sequences to attain more robust results.

Table 1 shows that the odometry head without the optical flow network can achieve remarkable inference speeds. The lightweight manner of the network can also be seen from the approximately 1GB of GPU VRAM it uses without the optical flow part. With online optical flow calculations, memory usage rises to 1.6 GB which is also tolerable. The more critical issue in this case is the speed drop which can later be mended with a more lightweight flow estimator than GMA.

## 5 Discussion

We have introduced a novel framework for all through Deep Learning based visual SLAM calculation. Our results have shown that our method is capable of producing results that are en par with the SOTA. For that, we used our state machine-like architecture shown in Fig. 1.

An important feature of our solution is that we used transfer learning based on a pretrained modern optical-flow estimator, namely GMA. Thanks to its modularity, our solution can be modified to make use of any other optical flow estimator solution. In our first experiments, we calculated the optical flow fields offline, and during inference we used the pre-calculated estimates as inputs. End-to-end training of the VO portion could yield better results, while resulting in longer execution times – we have experimented with this solution briefly, but more testing is still required.

It is important to note that the current solution does not contain loop closure detection, which is an important element of the mapping component of SLAM algorithms. We expect significant efficiency improvement from introducing this addition.

As mentioned previously, the results for our solution are shown in Table 1 were obtained from the KITTI dataset website. In our experiments, we have only used sequence 00 of the KITTI dataset. It is important to note that this might result in some generality limitations. A proof of this assumption is that the qualitative evaluation of running our algorithm on other sequences resulted in poor or hardly recognizable trajectories. In our future work, we plan to put more emphasis on generalization as well and evaluate our proposed solution on other sequences of the KITTI benchmark as well as other datasets.

Future development opportunities also include the parallelization of the separately executed sections of our framework. Through the invocation of several GPUs, the execution time of our solution could be significantly decreased while leaving the efficiency and accuracy of the results unaffected.

Another development possibility is to use the same data sequence for odometry training but with reversed order so the network could learn backward movements as well. This could be favorable for such an optimization process wherein first the forward propagation is done on a sequence, then the same sequence is used to calculate the odometry in reverse mode and the average of the two estimates (the backward converted to a forward estimate) is used.

The method introduced here is one of the key building blocks of the CrowdMapping framework presented in [5]. The integration of ATDN vSLAM with the autonomous traffic-aiding database-creation framework is in progress.

## 6 Conclusion

In this paper, we have introduced a novel, highly modular, all-through deep neural (ATDN) architecture for vSLAM. We have introduced the Embedding Distance Loss (EDL) function, which we used during the training process for the ATDN vSLAM architecture. We have presented the results of our system on the widely used KITTI dataset, where it achieved 4.4% translation and 0.0176 deg/m rotational error with an impressive 0.006 s runtime without and a still manageable 0.27 s runtime with online optical flow calculation. We have also shown the comparison between the results of our system and those of state-of-the-art solutions using the KITTI benchmark. We have identified key areas where our architecture can be further developed, including the lack of generalization for the entire KITTI dataset or other datasets, the introduction of loop closure for better mapping results, and module parallelization for better performance.

**Acknowledgement**
The research was supported by the Ministry of Innovation and Technology NRDI Office within the framework of the Artificial Intelligence National Laboratory Program.